\documentclass{article}

\PassOptionsToPackage{numbers}{natbib}
\usepackage[final]{neurips_2022_ml4ad}
\usepackage[utf8]{inputenc} 
\usepackage[T1]{fontenc}    
\usepackage{hyperref}       
\usepackage{url}            
\usepackage{booktabs}       
\usepackage{amsfonts}       
\usepackage{nicefrac}       
\usepackage{microtype}      
\usepackage{xcolor}         
\usepackage{float}
\usepackage{graphicx}
\usepackage{verbatim}       

\usepackage[utf8]{inputenc} 
\usepackage[T1]{fontenc}    
\usepackage{hyperref}       
\usepackage{url}            
\usepackage{booktabs}       
\usepackage{amsfonts}       
\usepackage{nicefrac}       
\usepackage{microtype}      
\usepackage{xcolor}         
\usepackage{graphicx}
\usepackage{amsmath, bm}
\usepackage{wrapfig,lipsum,booktabs}
\usepackage{tikz}
\usepackage{multirow}
\newcommand*\circled[1]{\tikz[baseline=(char.base)]{\node[shape=circle,draw,inner sep=0.8pt] (char) {#1};}}

\newcommand{\eg}{\textit{e}.\textit{g}.}
\newcommand{\etc}{\textit{etc}}
\newcommand{\ours}{Fast-BEV }

\title{Fast-BEV: Towards Real-time On-vehicle \\ Bird’s-Eye View Perception}

\author{
Bin Huang$^1$\thanks{Equal contribution.}
\And
Yangguang Li$^{1*}$
\And
Enze Xie$^{2*}$
\And
Feng Liang$^{3*}$
\And
Luya Wang$^4$
\And
Mingzhu Shen$^1$
\And
Fenggang Liu$^1$
\And
Tianqi Wang$^2$
\And
Ping Luo$^2$
\And
Jing Shao$^1$
\And
\texttt{$^1$ SenseTime, $^2$ The University of Hong Kong}\\
\texttt{$^3$ The University of Texas at Austin}\\
\texttt{$^4$ Beijing University of Posts and Telecommunications}\\
\mbox{\texttt{\{huangbin1,liufenggang,shaojing\}@senseauto.com}}\\
\mbox{\texttt{\{liyangguang,shenmingzhu\}@sensetime.com}}\\
\mbox{\texttt{\{xieenze,wangtq\}@connect.hku.hk}},
\mbox{\texttt{wangluya@bupt.edu.cn}}\\
\mbox{\texttt{jeffliang@utexas.edu}},
\mbox{\texttt{pluo@cs.hku.hk}}\\
}
\begin{document}

\maketitle

\begin{abstract}
Recently, the pure camera-based Bird’s-Eye-View (BEV) perception removes expensive Lidar sensors, making it a feasible solution for economical autonomous driving.
However, most existing BEV solutions either suffer from modest performance or require considerable resources to execute on-vehicle inference.
This paper proposes a simple yet effective framework, termed Fast-BEV, which is capable of performing real-time BEV perception on the on-vehicle chips.
Towards this goal, we first empirically find that the BEV representation can be sufficiently powerful without expensive view transformation or depth representation.
Starting from M$^2$BEV~\cite{xie2022m} baseline, we further introduce (1) a strong data augmentation strategy for both image and BEV space to avoid over-fitting (2) a multi-frame feature fusion mechanism to leverage the temporal information (3) an optimized deployment-friendly view transformation to speed up the inference.
Through experiments, we show \ours model family achieves considerable accuracy and efficiency on edge.
In particular, our M1 model (R18@256×704) can run over 50FPS on the Tesla T4 platform, with 47.0\% NDS on the nuScenes validation set.
Our largest model (R101@900x1600) establishes a new state-of-the-art 53.5\% NDS on the nuScenes validation set. The code is released at: \url{https://github.com/Sense-GVT/Fast-BEV}.

\end{abstract}

\section{Introduction} 
An accurate 3D perception system is essential for autonomous driving.
Classic methods~\cite{zhou2018voxelnet,lang2019pointpillars,shi2019pointrcnn} rely on the accurate 3D information provided by Lidar point clouds. 
However, Lidar sensors usually cost thousands of dollars~\cite{neuvition_2022}, hindering their applications on economical vehicles. 
Pure camera-based Bird’s-Eye-View (BEV) methods~\cite{xie2022m,li2022bevformer,huang2021bevdet,huang2022bevdet4d,li2022bevdepth} have recently shown great potential for their impressive 3D perception capability and economical cost.

To perform 3D perception from 2D image features, state-of-the-art BEV methods on nuScenes~\cite{caesar2020nuscenes} either uses implicit/explicit depth based projection~\cite{philion2020lift,huang2021bevdet,li2022bevdepth} or transformer based projection~\cite{chitta2021neat,li2022bevformer}.
However, they are difficult to deploy on on-vehicle chips: (1). Method with depth distribution prediction usually requires multi-thread CUDA kernels to speed up inference, which is inconvenient to operate on chips that are resource-constrained or not supported by inference libraries. (2). Attention mechanism within transformer needs dedicated chips to support.
Moreover, they are time-consuming in inference, which prevents them from actual deployment.
In this paper, we aim to design a BEV perception framework with leading performance, friendly deployment, and high inference speed for \emph{on-vehicle chips}.

\begin{figure}[t]
    \begin{center}
    \vspace{-1em}
    \includegraphics[width=0.98\textwidth]{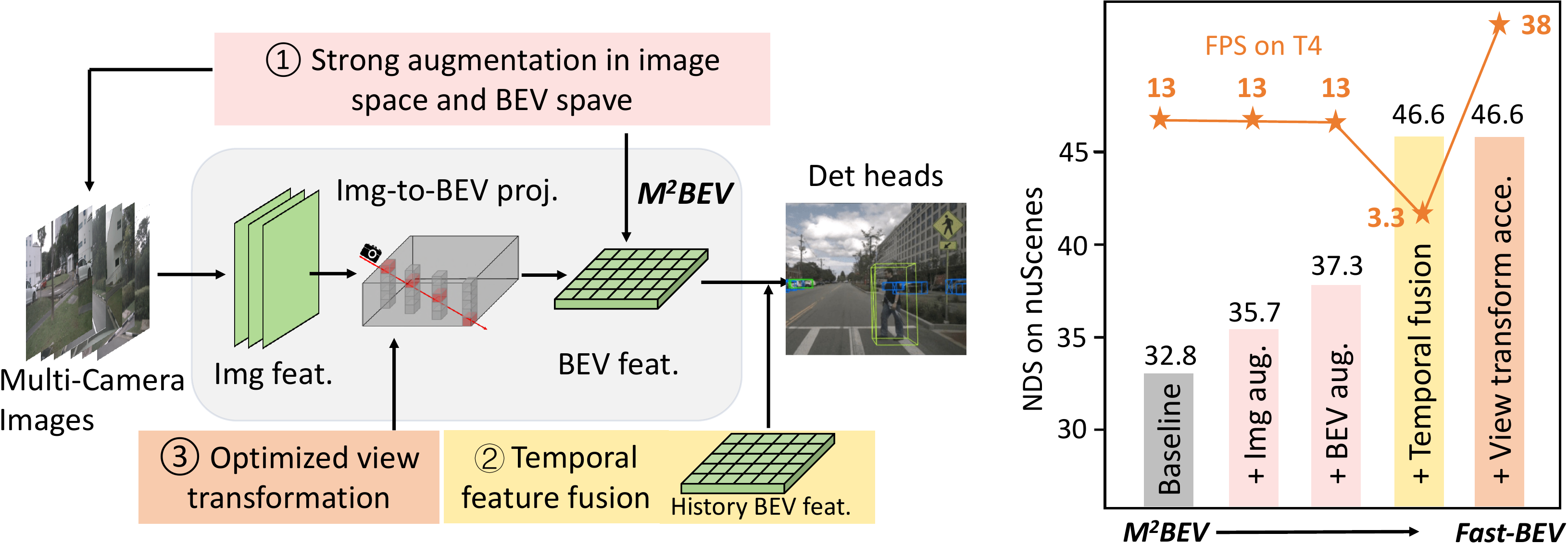}
    \end{center}
    \caption{Built upon M$^2$BEV, the proposed \ours first incorporates \protect\circled{1} strong augmentations to avoid over-fitting and \protect\circled{2} a multi-frame feature fusion mechanism to leverage the temporal information, leading to state-of-the-art performance.
    We further propose to optimize the \protect\circled{3} view transformation to be more deployment friendly for on-vehicle platforms.
    }
    \vspace{-1.5em}
\label{fig:fig_pipeline}
\end{figure}

Based on these observations, we choose to follow the principle of M$^2$BEV~\cite{xie2022m} which assumes a uniform depth distribution along the camera ray during image-to-BEV view transformation. 
We propose Fast-BEV, a stronger and faster fully convolutional BEV perception framework without expensive view transformer~\cite{wang2022detr3d,li2022bevformer} or depth representation~\cite{huang2021bevdet,huang2022bevdet4d,li2022bevdepth}.
We first verify the effectiveness of two training techniques by implementing them in the M$^2$BEV framework, namely, strong data augmentation~\cite{huang2021bevdet,li2022bevdepth} and temporal fusion~\cite{huang2022bevdet4d,li2022bevformer}. 
More specifically, additional data augmentation is applied on both image and BEV space to avoid over-fitting, which yields better performance.
We also extends the M$^2$BEV from spatial-only space to spatial-temporal space via introducing the temporal feature fusion module, enabling current key-frame leverage the information from history frames.
After integrating these two training techniques, our \ours can achieve comparable performance with state-of-the-art methods~\cite{huang2022bevdet4d,li2022bevdepth}.

\ours inherits the image-to-BEV transformation from M$^2$BEV. 
Although it is already fast, we empirically show that it can be further optimized for on-vehicle platforms.
We find that the projection from image space to voxel space dominates the latency.
Specifically, M$^2$BEV first computes the 2D-to-3D projection index for each camera view. 
After projection, it needs to aggregate the voxel features from all views.
We identify that each step can be significantly accelerated. 
First for the projection index, given the fact that camera positions and their intrinsic/extrinsic parameters are fixed when the perception system is built, we don't need to compute the \emph{same} index for each iteration.
Thus, we propose to pre-compute the fixed projection indexes and store them as a static look-up-table, which is super efficient during inference.
Second for the voxel aggregation, we observe the voxel features are highly sparse, \eg, only about 17\% positions are non-zero for 6-view dataset.
This is because each voxel feature only has the information of one camera.
We propose to generate a dense voxel feature to avoid the expensive voxel aggregation.
Specifically, we let image features from all camera views project to the \emph{same} voxel feature (see Section.~\ref{sec:acc}).
Combining these two acceleration designs improves the speed of the original M$^2$BEV by an order of magnitude. 

The proposed Fast-BEV model family shows great performance and can be easily deployed on on-vehicle platforms.
On the nuScenes~\cite{caesar2020nuscenes} dataset, our M1 model (R18@256×704) can run over 50FPS on the Tesla T4 platform, with considerable 46.9\% NDS performance.
Our largest model (R101@900x1600) establishes a new state-of-the-art 53.5\% NDS on the nuScenes validation set.
In conclusion, our contributions are summarized as follows:

\begin{itemize}

\item  We verify the effectiveness of two techniques: strong data augmentation and multi-frame temporal fusion on M$^2$BEV, enabling \ours achieve the SOTA performance.
\item We propose two acceleration designs: pre-computing the projection index and projecting to the same voxel feature, making \ours to be easily deployed on the on-vehicle chips with fast inference speed.

\item To the best of our knowledge, the proposed \ours is the first deployment-oriented work targeted on the challenging real-time on-vehicle BEV perception. We hope our work can shed light on the industrial-level, real-time, on-vehicle BEV perception.

\end{itemize}

\section{Related Work}
\noindent \textbf{Monocular 3D Perception.}
Compared to lidar, cheaper cameras provide richer semantic information and are immune to weather conditions.
Camera-based 3D object detection, given only image input, aims to estimate the object category, location, dimension, and orientation in the 3D space. 
A practical approach in monocular 3D detection is to predict 3D bounding boxes based on 2D image features. 
M3D-RPN~\cite{brazil2019m3d} proposes a 3D region proposal network and depth-aware convolutional layers to improve 3D scene understanding. 
Following FCOS~\cite{tian2019fcos}, FCOS3D~\cite{wang2021fcos3d} directly predicts 3D bounding boxes for each object by converting 3D targets into the image domain.
Further, PGD~\cite{wang2022probabilistic} uses the relations across the objects and a probabilistic representation to capture depth uncertainty to facilitate depth estimation for 3D object detection. 
DD3D~\cite{park2021pseudo} benefits from depth pre-training and significantly improve end-to-end 3D detection.
Besides object detection, another main perception task in autonomous driving is semantic segmentation in BEV for the targets to vectorially restore the surrounding environment. 
Some recent works~\cite{chitta2021neat,gosala2022bird,pan2020cross,roddick2020predicting} take 2D features into a 3D BEV representation based on a single view and perform BEV segmentation.

\smallskip

\noindent \textbf{Surrounding 3D Perception.}
The monocular 3D perception benchmark~\cite{geiger2012we} with only single-view images is insufficient for complicated tasks. 
Recently, some large-scale benchmarks~\cite{caesar2020nuscenes,sun2020scalability} have been proposed with much data and surrounding views, further promoting the field of 3D perception. 
The essential to 3D object detection is that all predictions correspond to the ground truth in 3D space.
One typical approach is to predict the depth of the camera-based input and convert the depth s into a pseudo-LiDAR to mimic the lidar signal ~\cite{wang2019pseudo,you2019pseudo,qian2020end}. 
Recently there has been another surge of interest in transforming image features to stereo representation like BEV(Bird’s-Eye-View) and 3D voxels~\cite{rukhovich2022imvoxelnet}. ~\cite{wang2019pseudo} transforms Pseudo-LiDAR obtained by visual depth estimation into BEV features.  
CaDDN ~\cite{reading2021categorical} uses a predicted categorical depth distribution for each pixel to project contextual features to BEV.
LSS ~\cite{philion2020lift} explicitly predicts depth distribution with a proposed view transform and projects image features onto BEV.
OFT~\cite{roddick2018orthographic} and ImVoxelNet~\cite{rukhovich2022imvoxelnet} generate the voxel representation of the scene by projecting the pre-defined voxels onto image features.
With a similar 4-stage framework~\cite{pan2020cross,roddick2020predicting,yang2021projecting}, BEVDet~\cite{huang2021bevdet} and M$^2$BEV~\cite{xie2022m} effectively extend CaDDN ~\cite{reading2021categorical} and OFT ~\cite{roddick2018orthographic} to multi-camera 3D object detection. 
Another class of methods is based on explicit pre-defined grid-shaped BEV queries. 
BEVFormerr ~\cite{li2022bevformer} performs 2D-to-3D transformation based on spatial cross-attention.  
Following DETR~\cite{carion2020end}, DETR3D ~\cite{wang2022detr3d} and Graph-DETR3D~\cite{chen2022graph} generates 3D reference points from BEV queries to sample the correlative 2D features. 
PETR ~\cite{liu2022petr} further introduces the 3D coordinate generation to perceive the 3D position-aware features and avoid generating 3D reference points.

\noindent \textbf{Multi-frame Fusion / Temporal Alignment.}
Lidar-based detectors can easily adopt multi-frame fusion ~\cite{yin2021center,lang2019pointpillars,yan2018second} to improve the velocity estimation accuracy and achieve better 3D object detection.
In contrast, the existing pure vision paradigms perform relatively poorly in predicting time-relevant targets due to the inability to access time cues. 
As an intermediate feature that simultaneously combines visual information from multiple multi-cameras at one time, BEV is suitable for temporal alignment.
~\cite{saha2021translating} proposes the Dynamics Module to use past spatial BEV features to learn a spatial-temporal BEV representation.
BEVFormer ~\cite{li2022bevformer} proposes a temporal self-attention to fuse the history BEV information recurrently, similar to the hidden state of RNN models.
BEVDet4D ~\cite{huang2021bevdet} extends the BEVDet ~\cite{huang2021bevdet} by aligning the multi-frame features and exploiting spatial correlations in ego-motion.
PETRv2 ~\cite{liu2022petrv2} extends the temporal version from PETR ~\cite{liu2022petr} to directly achieve temporal alignment in 3D space based on the perspective of 3D position embeddings.

\begin{figure}[t]
    \centering
    \includegraphics[width=1\textwidth]{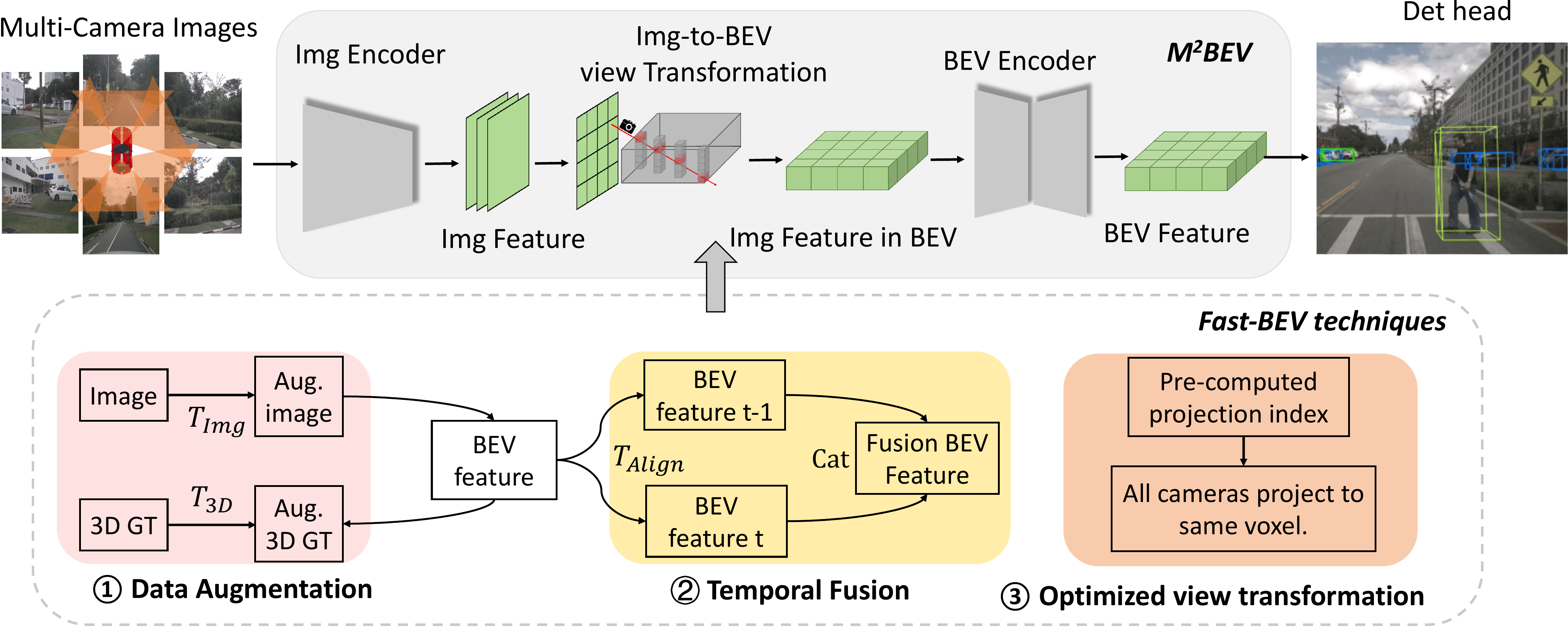}
    \caption{Overview of Fast-BEV. Same as M$^2$BEV, the multi-camera images are first fed into a image encode to extract image features. Then, image features are transformed into BEV space via view transform.
    A BEV encoder and task-specific heads are followed to perform perception tasks.
    The proposed \ours further applies \protect\circled{1} data augmentation on image and BEV domain, \protect\circled{2} temporal multi-frame fusion on M$^2$BEV. We further propose to pre-compute the image-to-voxel index and let all cameras project to same dense voxel to make  \protect\circled{3} view transformation model deployment friendly.}
\label{fig:method_overview}
\end{figure}

\section{Methods}
\subsection{Revisiting M$^2$BEV}

M$^2$BEV is one of the first works to solve the \textbf{M}ulti-camera \textbf{M}ulti-task perception with unified \textbf{BEV} representation. 
It is also more applicable for on-vehicle platforms since it doesn't have expensive view transformer or depth representation.
As denoted in the top part of Figure~\ref{fig:method_overview}, the input of M$^2$BEV is multi-camera RGB images, and the output is the predicted 3D bounding box (including velocity) and map segmentation results. 
M$^2$BEV has four key modules: (1) A 2D image encoder that extracts image features from multi-camera images (2) An image-to-BEV (2D$\rightarrow$3D) view transformation module that maps the 2D image feature into 3D BEV space (3) A 3D BEV encoder that processes the 3D features and (4) Task-specific heads that perform perception tasks, \eg, 3D detection.

\subsection{Overall Architecture of \ours}

Although M$^2$BEV can achieve competitive results, we find its performance and efficiency can be further improved. 
As denoted in the bottom part of Figure~\ref{fig:method_overview}, we integrates three techniques into M$^2$BEV, leading to our stronger and faster Fast-BEV.

\protect\circled{1} Data augmentation. We empirically observe that severe over-fitting problem happened during the later training epochs in M$^2$BEV. 
This is because no data augmentation is used in original M$^2$BEV.
Motivated by recent work~\cite{huang2021bevdet,li2022bevdepth}, we add strong 3D augmentation on both image and BEV space, such as random flip, rotation \etc. More details are in Section~\ref{sec:aug}

\protect\circled{2} Temporal fusion. In real autonomous driving scenarios, the input is temporally continuous and has tremendous complementary information across time. 
For instance, one pedestrian partially occluded at current frame might be fully visible in the past several frames.
Thus, we extend M$^2$BEV from spatial-only space to spatial-temporal space via introducing the temporal feature fusion module, similar with ~\cite{huang2022bevdet4d,li2022bevformer}.
More specifically, we use current frame BEV features and stored history frame features as input and train \ours in an end-to-end manner. 
More details are in Section~\ref{sec:multiframe}

\protect\circled{3} Optimized view transformation. 
We find that the projection from image space to voxel space dominates the latency.
We propose to optimize the projection from two perspectives: (1) we pre-compute the fixed projection indexes and store them as a static look-up-table, which is super efficient during inference. 
(2) We let all the cameras project to the same voxel to avoid expensive voxel aggregation. Our proposal is not like the improved view transform schemes~\cite{li2022bevdepth,huang2021bevdet,liu2022bevfusion} based on Lift-Splat-Shoot, which requires the development of cumbersome and difficult DSP/GPU parallel computing, it is fast enough to use only CPU computing, which is very convenient for deployment. More details are in Section~\ref{sec:acc}

We would like to clarify that \protect\circled{1} and \protect\circled{2} are inspired by the concurrent leading works~\cite{huang2022bevdet4d,li2022bevformer} and we do not intend to regard these two parts as novel designs. 
These improvements make our proposed pipeline, Fast-BEV, become a SOTA method while keeping its simplicity for on-vehicle platforms.

\begin{figure}[t]
    \centering
    \includegraphics[width=1\textwidth]{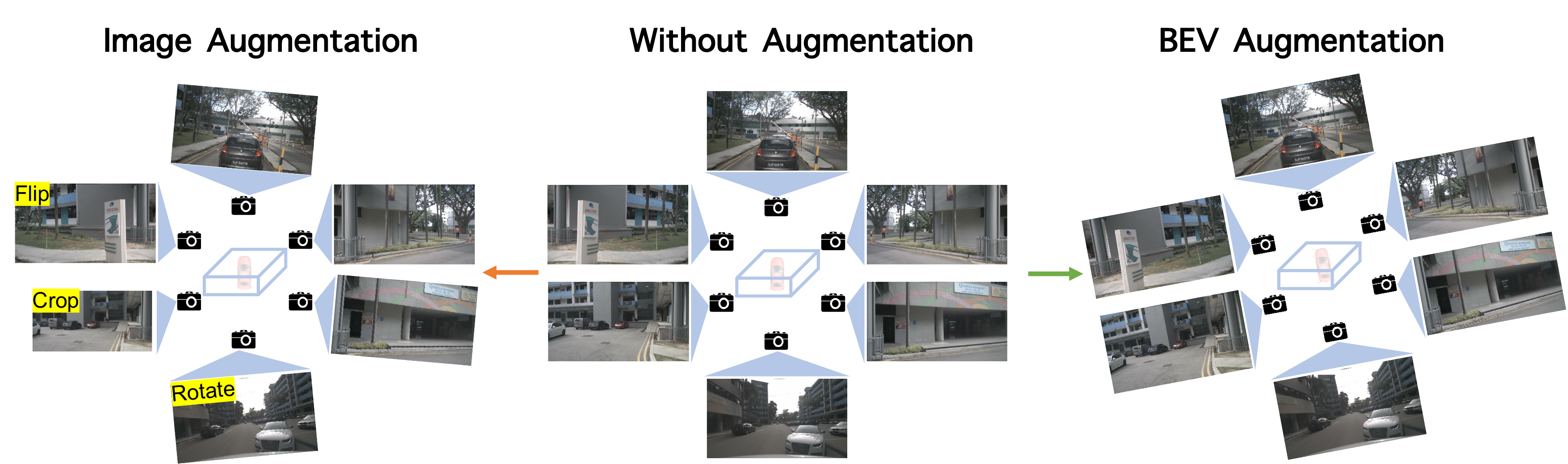}
    \caption{Examples of the data augmentation used in Fast-BEV. 
    The middle figure shows the original M$^2$BEV, which does not use data augmentation.
    The left figure shows the image augmentation and some augmentation types such as random flip, crop and rotate. 
    The right figure shows one type of BEV augmentation, random rotation.
    }
\label{fig:aug_fig}
\end{figure}

\subsection{Data Augmentation} \label{sec:aug}
We add data augmentations in both image space and BEV space, mainly following BEVDet~\cite{huang2021bevdet}. 

\textbf{Image Augmentation.} The data augmentation in 3D object detection is more challenging than that of 2D detection since the images in 3D scenarios have direct relationship with 3D camera coordinates. Thus, if we apply data augmentation on images, we need also change the camera intrinsic matrix~\cite{huang2021bevdet}.
For the augmentation operations, we basically follow the common operations, \eg, flipping, cropping and rotation.
In the left part of Fig.~\ref{fig:aug_fig}, we show some examples of image augmentations.

\smallskip
\textbf{BEV Augmentation.} 

\begin{wrapfigure}{r}{0.5\textwidth}
  \begin{center}
    \includegraphics[width=0.48\textwidth]{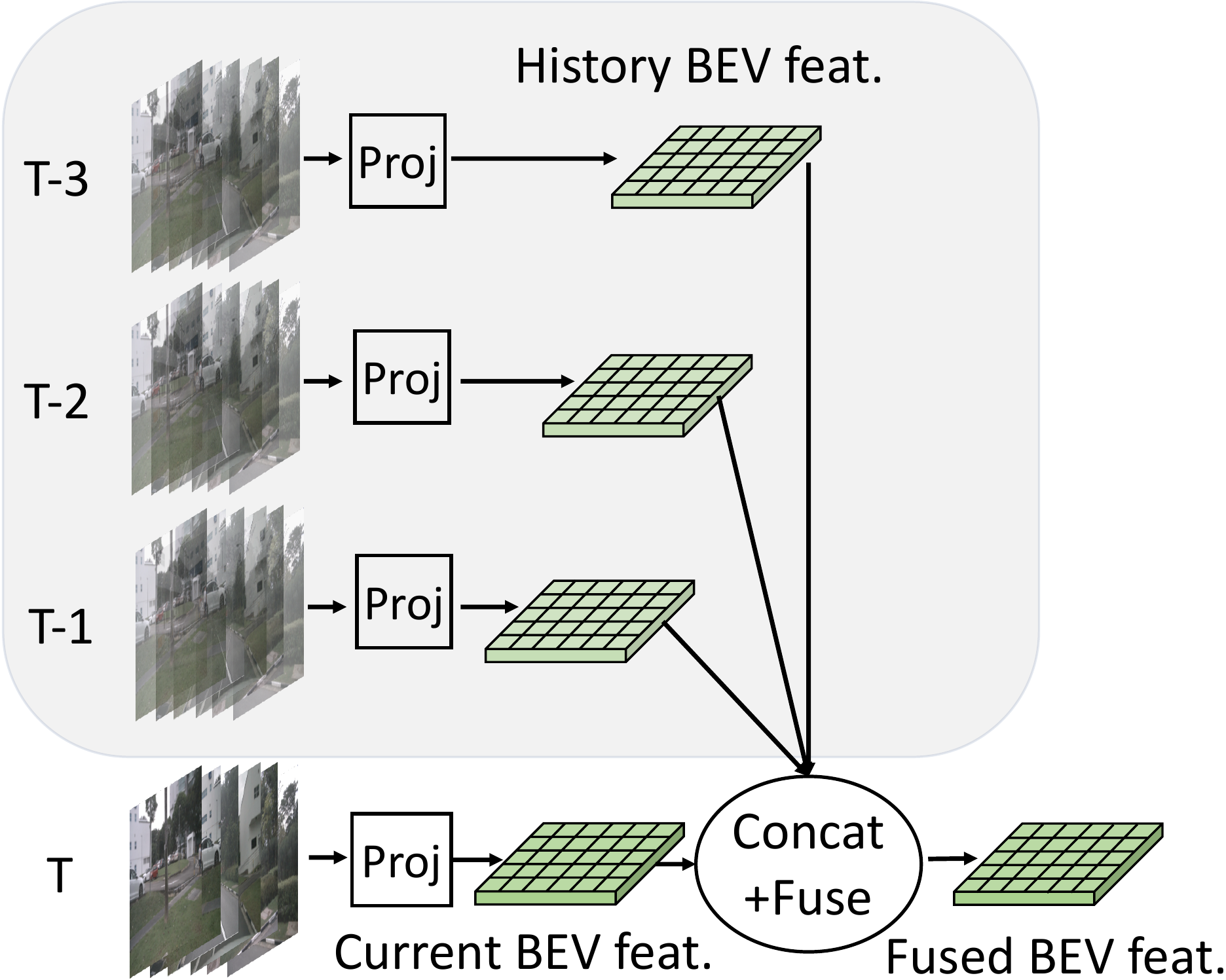}
  \end{center}
  \caption{Illustration of the temporal multi-frame feature fusion module. The three history frames are first extracted feature and projected to the respective BEV space, then aligned to the current frame with camera extrinsic and global coordinate. Finally, we directly concatenate these multi-frame BEV features in the channel dimension.}
  \vspace{-3em}
  \label{fig:fig_fuse}
\end{wrapfigure}
Similar to image augmentation, similar operations can be applied to the BEV space, such as flipping, scaling and rotation. Note that the augmentation transformation should be applied on both the BEV feature map and the 3D ground-truth box to keep consistency. The BEV augmentation transformation can be controlled by modifying the camera extrinsic matrix accordingly.
In the right part of Fig.~\ref{fig:aug_fig}, we show the random rotation augmentation, a type of BEV augmentation.

\subsection{Multi-Frame Feature Fusion} \label{sec:multiframe}

Inspired by BEVDet4D~\cite{huang2022bevdet4d} and BEVFormer~\cite{li2022bevformer}, we also introduce the history frame into the current frame for temporal feature fusion.
Here we sample the current frame with three history keyframes; each keyframe has a 0.5s interval.
We adopted the multi-frame feature alignment method from BEVDet4D. 
As shown in Fig.~\ref{fig:fig_fuse}, after we got four aligned BEV features, we directly concatenate them and feed them to the 3D encoder.
In the training phase, the history frame features are extracted online using the image encoder.
In the testing phase, the history frame feature can be saved offline and directly taken out for acceleration.

\smallskip

\textbf{Compare with BEVDet4D and BEVFormer.}
BEVDet4D only introduces one history frame, which we argue is insufficient to leverage history information. 
\ours uses three history frames, resulting in significant performance improvement. 
BEVFormer is slightly better than BEVDet4D by using two history frames. However, due to memory issues, in the training phase, the history feature is detached without grad, which is not optimal. Moreover, BEVFormer uses an RNN-style to sequentially fuse features, which is inefficient. 
In contrast, all the frame in \ours is trained in an end-to-end manner, which is more training friendly with common GPUs.

\subsection{Optimized view transformation} \label{sec:acc}

\begin{figure}[t]
    \centering
    \includegraphics[width=0.8\textwidth]{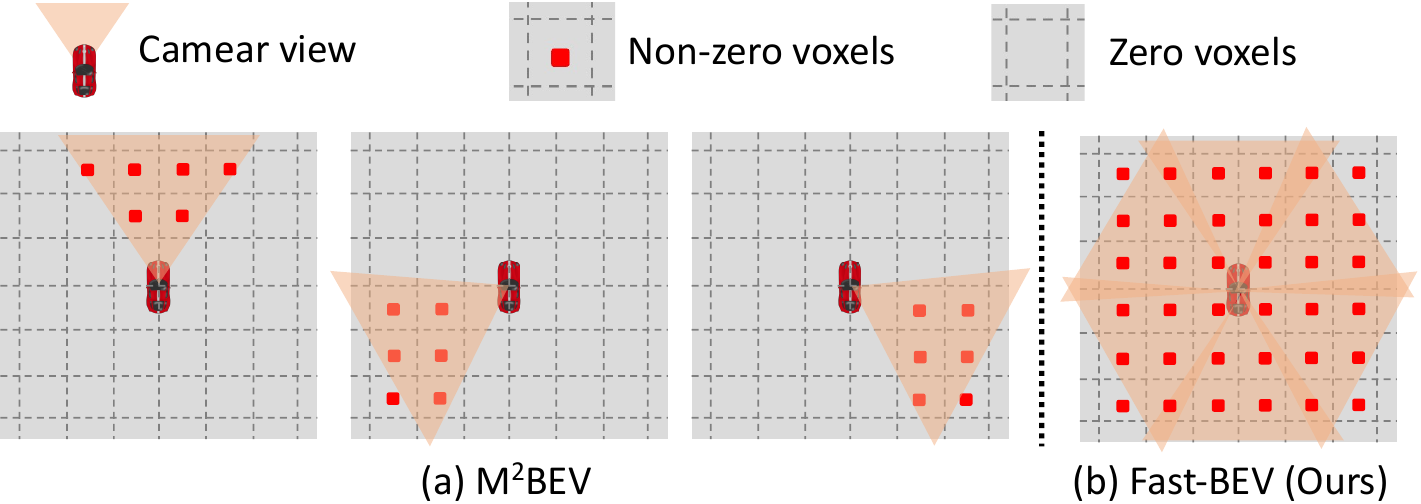}
    \caption{(a) In M$^2$BEV baseline, each camera has one sparse voxel (only $\sim$17\% positions are non-zeros). An expensive aggregation operation is needed to combine the sparse voxels. (b) The proposed Fast-BEV let all cameras project to one dense voxel, avoiding the expensive voxel aggregation.
    }
\label{fig:accelerator}
\end{figure}

View Transformation is the critical component to transform features from 2D image space to 3D BEV space, which typically takes much time in the whole pipeline. 
Lift-Splat-Shoot~\cite{philion2020lift} is a classic method for view transformation.
Although some acceleration techniques~\cite{li2022bevdepth,huang2021bevdet,liu2022bevfusion} have been proposed for advanced GPU devices~(\eg, NVIDIA Tesla A100, V100), but the optimization can not be easily transferred to other devices such as edge chips.
Another category of view transformation is M$^2$BEV~\cite{xie2022m} which assumes the depth distribution is uniform along the ray. 
The advantage is that once we get the intrinsic/extrinsic parameters of cameras, we can easily know the 2D to 3D projection.
Since no learnable parameters are used here, we can easily compute the corresponding matrix between points in the 2D feature maps and the BEV feature map. 
We follow the projection method of M$^2$BEV, and further accelerate it from two perspectives: pre-compute projection index and dense voxel feature generation.

The projection index is the mapping index from 2D image space to 3D voxel space. 
Because our method do not rely on the data-dependent depth prediction nor transformer, the projection index will be the \emph{same} for every input.
Thus, we can pre-compute the fixed projection index and store it.
During inference, we can get the projection index via querying the look-up-table, which is a very cheap operation on edge devices.
Moreover, if we extend from the single frame to multiple frames, we can also easily pre-compute the intrinsic and extrinsic parameters and pre-align them to the current frame.

\begin{wraptable}{rt}{0.57\textwidth}
    \centering
    \vspace{-20pt}
    \caption{Latency profiling for view transformation of different methods. The latency is tested on Tesla A100 GPU and Intel i7 CPU in milliseconds~(ms).}
	\label{table:view_transformation}
	\vspace{5pt}
    \resizebox{0.99\linewidth}{!}{
	\begin{tabular}{l|c|c|c}
    \hline\noalign{\smallskip}
    Method          & Projection      & GPU  Latency     & CPU Latency  \\
    \noalign{\smallskip} \hline \noalign{\smallskip}
    BEVDet~\cite{huang2021bevdet}    &Depth        & $\sim$125  & -       \\
    BEVDepth~\cite{li2022bevdepth}  &Depth & $\sim$1.2 & -      \\
    BEVFormer~\cite{li2022bevformer} &Transformer  & 25   & -      \\
    M$^2$BEV~\cite{xie2022m}  & None & - & 54       \\
    \noalign{\smallskip} \hline \noalign{\smallskip}
    \ours (Ours) & None & - & 0.8       \\
    \noalign{\smallskip} \hline \noalign{\smallskip}
    \end{tabular}
    }
    \vspace{-20pt}
\end{wraptable}

M$^2$BEV will store a voxel feature for each camera view and then aggregate them to generate the final voxel feature (see Figure~\ref{fig:accelerator}).
Because each camera only has limited view angle, each voxel feature is very sparse, \eg, only about 17\% positions are non-zeros.
We identify the aggregation of these voxel features is very expensive due to the their huge size.
We propose to generate a dense voxel feature to avoid the expensive voxel aggregation. 
Specifically, we let image features from all camera views project to the same voxel feature, leading to one dense voxel at the end.

In Table~\ref{table:view_transformation}, we profile the view transformation latency of four different methods.
We find that (1) BEVDepth~\cite{li2022bevdepth} achieves the best latency on GPU but it requires dedicated parallel computing support, making it not applicable to CPU.
(2) Compared with the M$^2$BEV baseline, the proposed Fast-BEV achieves orders of magnitude speedup on CPU.

\section{Experiments}

\subsection{Setup}
\textbf{Dataset Description}
We evaluate our \ours on nuScenes dataset~\cite{caesar2020nuscenes}, which contains 1000 autonomous driving scenes with 20 seconds per scene.
The dataset is split into 850 scenes for training/validation and the rest 150 for testing.  
While nuScenes dataset provides data from different sensors, we only use the camera data.
The cameras have six views: $\tt front\_left$, $\tt front$, $\tt front\_right$, $ \tt back\_left$, $\tt back$, $\tt back\_right$. 

\textbf{Evaluation metrics.} 
To comprehensively evaluate the detection task, we use the standard evaluation metrics of mean Average Precision (mAP), and nuScenes detection score (NDS) for 3D object detection evaluation.
In addition, in order to calculate the precision of the corresponding aspects (\textit{e.g.}, translation, scale, orientation, velocity, and attribute), we use the mean Average Translation Error (mATE), mean Average Scale Error (mASE), mean Average Orientation Error(mAOE), mean Average Velocity Error(mAVE), and mean Average Attribute Error(mAAE) as the metric.

\textbf{Implementation Details.}
For training, we use AdamW optimizer with learning rate $1e^{-3}$ and the weight decay is set to $1e^{-2}$.
``Polylr'' scheduler is adopted to gradually decrease the learning rate. 
We also use ``warmup'' strategy for the first 1000 iterations.
For data augmentation hyper-parameters, we basically follow BEVDet~\cite{huang2021bevdet}.
Without specific notification, all models are trained on 32 A100 GPUs for 48 epochs.
For comparison with state-of-the-art, we train 20 epochs with CBGS~\cite{zhu2019class}. 
We mainly use ResNet-50 as the backbone for the ablation study to verify the ideas quickly. 
For on-vehicle inference speed test, we test with one sample per batch which contains 6 view images.

\subsection{Compare with state-of-the-art methods}

\setlength{\tabcolsep}{1.5pt} 
\begin{table}[t]
\centering
\caption{Comparison on the nuScenes \emph{val} set. ``L'' denotes LiDAR, ``C'' denotes camera and ``D'' denotes Depth/LiDAR supervision.
``\P'' indicates our method with scale NMS and test-time augmentation.
}
\vspace{-2mm}
\resizebox{\textwidth}{!}{
\setlength{\tabcolsep}{1.5pt}
\begin{tabular}{l|c|c|cccccc|c}
\hline\noalign{\smallskip}
Methods & Image Res. &  Modality & mAP$\uparrow$ & mATE$\downarrow$ &mASE$\downarrow$ &mAOE$\downarrow$ &mAVE$\downarrow$ &mAAE$\downarrow$ & NDS$\uparrow$\\
\noalign{\smallskip}
\hline
\noalign{\smallskip}
CenterPoint-Voxel~\cite{yin2021center}  &         -      & L        & 0.564 &   -   &    -   &     -    &     -  &    -   & 0.648  \\
CenterPoint-Pillar~\cite{yin2021center} &       -        & L        & 0.503 &   -    &    -   &     -    &  -     &     -  & 0.602  \\ 

\noalign{\smallskip}
\hline
\noalign{\smallskip}
\noalign{\smallskip}
\hline
\noalign{\smallskip}
FCOS3D~\cite{wang2021fcos3d}             & 900$\times$1600      & C        & 0.295 & 0.806 & 0.268 & 0.511   & 1.315 & 0.170 & 0.372  \\
BEVDet-R50~\cite{huang2021bevdet}         & 256$\times$704       & C        & 0.286 & 0.724 & 0.278 & 0.590   & 0.873 & 0.247 & 0.372 \\
PETR-R50~\cite{liu2022petr}           & 384$\times$1056      & C        & 0.313 & 0.768 & 0.278 & 0.564   & 0.923 & 0.225 & 0.381  \\
PETR-Tiny~\cite{liu2022petr} & 512$\times$1408      & C        & 0.361 & 0.732 & 0.273 & 0.497   & 0.808 & 0.185 & 0.431  \\
BEVDet4D-Tiny~\cite{huang2022bevdet4d}      & 256$\times$704       & C        & 0.323 & 0.674 & 0.272 & 0.503   & 0.429 & 0.208 & 0.453  \\
BEVDepth-R50~\cite{li2022bevdepth}        & 256$\times$704       & C\&D        & 0.351 & 0.639 & 0.267 & 0.479   & 0.428 & 0.198 & 0.475  \\
\noalign{\smallskip}
\hline
\noalign{\smallskip}
\textbf{Fast-BEV(R50)}  & 256$\times$704& C &0.334 & 0.665 &0.285 & 0.393   & 0.388 & 0.210 &0.473\\
\textbf{Fast-BEV(R50)\P}  & 256$\times$704& C & 0.346 & 0.667 &0.285 & 0.401   & 0.393 & 0.208 & \textbf{0.477}\\
\hline
FCOS3D-R101$\dagger$~\cite{wang2021fcos3d}   & 900$\times$1600& C        & 0.321 & 0.754 & 0.260 & 0.486   & 1.331 & 0.158 & 0.395  \\
DETR3D-R101$\dagger$~\cite{wang2022detr3d}    & 900$\times$1600 & C        & 0.347 & 0.765 & 0.267 & 0.392   & 0.876 & 0.211 & 0.422  \\
Ego3RT-V2-99$\dagger$~\cite{lu2022learning}   & 900$\times$1600 & C    & 0.478 & 0.582 & 0.272 & 0.316   & 0.683 & 0.202 & 0.534  \\
M$^2$BEV-X101$\dagger$~\cite{xie2022m}  & 900$\times$1600 & C    & 0.417 & 0.647 & 0.275 & 0.377   & 0.834 & 0.245 & 0.470  \\
PolarFormer-T-R101$\dagger$~\cite{jiang2022polarformer}   & 900$\times$1600  & C    & 0.432 & 0.648 & 0.270 & 0.348   & 0.409 & 0.201 & 0.528  \\
PETRv2-VoVNet-99$\dagger$~\cite{liu2022petrv2}  & 320$\times$800     & C    & 0.401 & 0.745 & 0.268 & 0.448   & 0.394 & 0.184 & 0.496  \\
DETR3D~\cite{wang2022detr3d}             & 900$\times$1600      & C        & 0.303 & 0.860 & 0.278 & 0.437   & 0.967 & 0.235 & 0.374  \\
BEVDet-Base ~\cite{huang2021bevdet}       & 512$\times$1408      & C        & 0.349 & 0.637 & 0.269 & 0.490   & 0.914 & 0.268 & 0.417 \\
PETR-R101 ~\cite{liu2022petr}         & 512$\times$1408      & C        & 0.357 & 0.710 & 0.270 & 0.490   & 0.885 & 0.224 & 0.421  \\
BEVDet4D-Base ~\cite{huang2022bevdet4d}     & 640$\times$1600      & C        & 0.396 & 0.619 & 0.260 & 0.361   & 0.399 & 0.189 & 0.515  \\
BEVFormer-R101~\cite{li2022bevformer}        &        900$\times$1600       & C        & 0.416 & 0.673 & 0.274 & 0.372   & 0.394 & 0.198 & 0.517  \\
BEVDepth-R101 ~\cite{li2022bevdepth}      & 512$\times$1408      & C\&D        & 0.412 & 0.565 & 0.266 & 0.358   & 0.331 & 0.190 & \textbf{0.535}  \\
\noalign{\smallskip}
\hline
\noalign{\smallskip}
\textbf{Fast-BEV(R101)} & 900$\times$1600& C & 0.402 & 0.582 &0.278 &0.304   & 0.328 & 0.209 & 0.531\\
\textbf{Fast-BEV(R101)\P} & 900$\times$1600& C & 0.413 & 0.584 &0.279 & 0.311   & 0.329 & 0.206 & \textbf{0.535}\\
\hline
\end{tabular}}
\label{tab:val}
\end{table}
\setlength{\tabcolsep}{1.4pt}

\begin{table}[t]
\begin{center}
\caption{The modular design of the serial models denoted as M0-4 are presented. The type of 2d encoder is chosen from R18/R34/R50. Voxel resolution is denoted as x-y-z for the view transformation from 2d features to BEV features. The number of block x and channel y denoted as xb-yc are clarified for 3D encoders.
The latency on T4 platform are evaluated with the sum of 3 parts including a 2D encoder, view transformation and a 3D encoder (from left to right in the latency breakdown column).
}
\label{table:serial_models}
\setlength{\tabcolsep}{3pt}
\scalebox{0.88}{
\begin{tabular}{c|c|c|c|c|c|c|c|c}
\noalign{\smallskip}\hline\noalign{\smallskip}
Name & 2D Encoder & Image Res.& Voxel Res.&3D Encoder & mAP & NDS& Latency (ms) & Latency breakdown\\
\noalign{\smallskip}\hline\noalign{\smallskip}
M0 & R18 & 256$\times$704 & 200$\times$200$\times$4 & 2b-192c & 0.284 & 0.427 & 14.5 & 4.6 / 2 / 7.9     \\
M1 & R34 & 256$\times$704 & 200$\times$200$\times$4 & 4b-224c & 0.326 & 0.470 & 16.9 & 5.8 / 2 / 9.1     \\
M2 & R34 & 320$\times$880 & 250$\times$250$\times$6 & 4b-224c & 0.332 & 0.472 & 35.2 & 8.6 / 4.8 / 21.8  \\
M3 & R50 & 320$\times$880 & 250$\times$250$\times$6 & 6b-256c & 0.346 & 0.482 & 39.3 & 10.8 / 4.8 / 23.7 \\
M4 & R50 & 384$\times$1056 & 300$\times$300$\times$6 & 6b-256c & 0.349 & 0.488 & 57.2 & 14.6 / 7.6 / 35.0 \\
\noalign{\smallskip}\hline\noalign{\smallskip}
\end{tabular}
}
\end{center}
\end{table}

We comprehensively compare the proposed Fast-BEV with the baseline method M$^2$BEV~\cite{xie2022m} and other recent methods like FCOS3D~\cite{wang2021fcos3d}, PETR~\cite{liu2022petr}, BEVDet~\cite{huang2021bevdet}, DETR3d~\cite{wang2022detr3d}, BEVDet4D~\cite{huang2022bevdet4d}, BEVFormer~\cite{li2022bevformer} and BEVDepth~\cite{li2022bevdepth} on nuScenes val set.

As shown in Table~\ref{tab:val}, Fast-BEV shows superior performance in mAP and NDS compare with existing advanced methods. 
For example, with ResNet-50 as the backbone, Fast-BEV achieves 0.346 mAP and 0.477 NDS, significantly outperforms BEVDet4D-Tiny~\cite{huang2022bevdet4d} (0.323 mAP and 0.453 NDS).
The model also exceeds other methods with a larger input resolution such as PETR-R50~\cite{liu2022petr} (0.313 mAP and 0.381 NDS).
Moreover, 
with the larger backbone ResNet-101 and  higher image resolution, Fast-BEV establish a new state-of-the-art 0.535 NDS, exceeding BEVDet4D-Base~\cite{huang2022bevdet4d} 0.515 NDS and BEVFormer-R101~\cite{li2022bevformer} 0.517 NDS without depth/lidar supervision.

\smallskip

\noindent{\textbf{Efficient Model Series.}} In order to fit different deployment scenarios, we design a series of efficient models from M0 to M4 as shown in Table~\ref{table:serial_models}. 
We set different 2D encoder, image resolution, voxel resolution and 3D encoder to design the model sizes.
Our M1 model (R18@256×704) can run over 50FPS on the Tesla T4 platform, with 0.470 NDS on the nuScenes validation set.

We also breakdown the latency into three parts in the last column of Table~\ref{table:serial_models}. 
As the 2D encoder grow larger from R18 to R50, the image resolution increase from $250\times704$ to $384\times1056$, the latency of this part increases by nearly 3 times. At the same time, when the voxel resolution increases from $200\times200\times4$ to $300\times300\times6$, the 3D encoder grows larger from 2b-192c to 6b-256c, the latency of view transformation and 3D encoder increases dramatically by nearly 4 times and 5 times, respectively.

\subsection{Detailed Analysis}
Unless explicitly stated, Fast-BEV analysis experiments for the detection task in this section use the ResNet-50 backbone and 2 frames to train 48 epochs. And the image resolution and voxel resolution are $250\times704$ and $200\times200\times6$, respectively.

\noindent{\textbf{Augmentation.}} As shown in Table~\ref{table:augmentation}, We observe that the performance is significantly improved whether using image augmentation or BEV augmentation alone. 
For image augmentation, mAP and NDS increased by 3.8\% and 2.8\%, respectively. 
For BEV augmentation, mAP and NDS increased by 1.8\% and 2.3\%, respectively. When the two augmentation are used together, the mAP and NDS can be further improved by 4.6\% and 4.4\%. 

\smallskip
\noindent{\textbf{Multi-Frame Feature Fusion.}} 
To investigate the effectiveness of multi-frame feature fusion, we show an ablation study in Table~\ref{table:sequantial}. 
When adding one history frame, the mAP and NDS are significantly improved with 2.8\% and 7.8\%. When further increase to four history frames, the mAP and NDS continue to improve by  with 3.0\% and 9.3\%, showing that temporal information is important for 3D detection.
\smallskip

\begin{minipage}{\textwidth}
\begin{minipage}[t]{0.31\columnwidth}
\makeatletter\def\@captype{table}
\caption{Ablation study of different augmentations with single frame.}
\centering
\begin{tabular}{ccc}
    \noalign{\smallskip}\hline\noalign{\smallskip}
    Method  & mAP & NDS\\
    \noalign{\smallskip}\hline\noalign{\smallskip}
    Baseline&  0.247 & 0.329\\
    +ImgAug &  0.285 & 0.357 \\
    +BEVAug &  0.265 & 0.352 \\
    +ImgBEVAug &0.293 & 0.373 \\
    \noalign{\smallskip}\hline\noalign{\smallskip}
\end{tabular}
\label{table:augmentation}
\end{minipage}
\hspace{2mm}
\begin{minipage}[t]{0.31\columnwidth}
\makeatletter\def\@captype{table}
\caption{Ablation study of sequential feature fusion from single frame to 4 frames. }
\centering
\begin{tabular}{ccc}
    \noalign{\smallskip}\hline\noalign{\smallskip}
    Method & mAP & NDS\\
    \noalign{\smallskip}\hline\noalign{\smallskip}
    1F &  0.293 & 0.373 \\
    2F & 0.321 & 0.451 \\
    4F & 0.323 & 0.466 \\
    \noalign{\smallskip}\hline\noalign{\smallskip}
\end{tabular}
\label{table:sequantial}
\end{minipage}
\hspace{2mm}
\begin{minipage}[t]{0.31\columnwidth}
\makeatletter\def\@captype{table}
\caption{Ablation study of voxel resolution. The resolution of the image is 384$\times$1056.}
\centering
\begin{tabular}{lcc}
    \noalign{\smallskip}\hline\noalign{\smallskip}
    Voxel Res.  & mAP & NDS\\
    \noalign{\smallskip}\hline\noalign{\smallskip}
    200$\times$200$\times$6& 0.352 & 0.476 \\
    200$\times$200$\times$12& 0.350 & 0.474 \\
    400$\times$400$\times$6& 0.337 & 0.467 \\
    400$\times$400$\times$12& 0.345 & 0.476 \\
    \noalign{\smallskip}\hline\noalign{\smallskip}
\end{tabular}
\label{table:v_input_size}
\end{minipage}
\end{minipage}

\begin{minipage}{\textwidth}
\begin{minipage}[t]{0.31\columnwidth}
\makeatletter\def\@captype{table}
\caption{Ablation study of image resolution.}
\centering
\begin{tabular}{lcc}
    \noalign{\smallskip}\hline\noalign{\smallskip}
    Image Res.  & mAP & NDS\\
    \noalign{\smallskip}\hline\noalign{\smallskip}
256$\times$448  &0.280 & 0.419 \\
256$\times$704  &0.321 & 0.451 \\
464$\times$800  &0.342 & 0.466 \\
544$\times$960  &0.345 & 0.472 \\
704$\times$1208 &0.358 & 0.478 \\
832$\times$1440 &0.368 & 0.491 \\
928$\times$1600 &0.369 & 0.488 \\
    \noalign{\smallskip}\hline\noalign{\smallskip}
\end{tabular}
\label{table:s_input_size}
\end{minipage}
\hspace{2mm}
\begin{minipage}[t]{0.31\columnwidth}
\makeatletter\def\@captype{table}
\caption{Ablation study of baseline and Fast-BEV epochs.}
\label{table:epoch}
\centering
\begin{tabular}{lccc}
    \noalign{\smallskip}\hline\noalign{\smallskip}
Method& Epochs & mAP & NDS\\
    \noalign{\smallskip}\hline\noalign{\smallskip}
Baseline &12 & 0.258 & 0.330\\
&24 & 0.257 & 0.338\\
&36 & 0.248 & 0.320\\
&48 & 0.247 & 0.327\\
    \noalign{\smallskip}\hline\noalign{\smallskip}
Fast-BEV&12&0.273 & 0.381 \\
&24&0.294 & 0.424 \\
&36&0.310 & 0.440 \\
&48&0.321 & 0.451 \\
    \noalign{\smallskip}\hline\noalign{\smallskip}
\end{tabular}
\end{minipage}
\hspace{2mm}
\begin{minipage}[t]{0.31\columnwidth}
\makeatletter\def\@captype{table}
\caption{Ablation study of different 2D and 3D encoders with 4 frames and 20 epochs with CBGS.}
\centering
\begin{tabular}{lccc}
        \noalign{\smallskip}\hline\noalign{\smallskip}
Encoder&Type& mAP & NDS\\
    \noalign{\smallskip}\hline\noalign{\smallskip}
2D&R18& 0.293 & 0.437 \\
&R50& 0.335 & 0.473 \\
&R101&0.345 & 0.482 \\
    \noalign{\smallskip}\hline\noalign{\smallskip}
3D&2$\times$Block   &0.320&0.446 \\
&4$\times$Block   &0.315&0.448 \\
&6$\times$Block   &0.321&0.451 \\
    \noalign{\smallskip}\hline\noalign{\smallskip}
\end{tabular}
\label{table:encoder}
\end{minipage}

\end{minipage}

\noindent\textbf{Resolution.} To investigate the effect of different resolutions of the input image and voxel, we perform an ablation study in Table~\ref{table:s_input_size} and Table~\ref{table:v_input_size}.
We first fix the voxel resolution to 200$\times$200$\times$6, and discretely take different image resolutions from 256$\times$448 to 928$\times$1600 for verification.
The results in Table~\ref{table:s_input_size} show that the increasing in resolution greatly helps to improve the performance of the model, and Fast-BEV achieves the best 49.1\% NDS with 832$\times$1440 input image size.

We then fix the image resolution to 384$\times$1056, and try to use different voxel resolutions as shown in Table~\ref{table:v_input_size}. 
We observe that $200\times 200\times 6$ works well for 3D detection, and increase the resolution from the spatial plane or the height dimension does not help improving the performance.

\smallskip

\noindent\textbf{2D/3D Encoder.} To evaluate the performance of different 2D encoders, we perform an ablation study in Table~\ref{table:encoder} (upper). As the encoder grows larger from ResNet-18 to ResNet-101, the mAP and NDS increases by a large margin with over 5\% and 4\%. 
In terms of 3D encoders, as shown in Table~\ref{table:encoder} (lower), when the encoder grows larger from 2 blocks to 6 blocks, the mAP and NDS for detection task increases by 0.1\% and 0.5\% respectively. The scale of the 2D encoder has a greater impact on performance than the 3D encoder.

\noindent{\textbf{Epoch.}} To investigate the influence of training epochs, we do an ablation study in Table~\ref{table:epoch}.
We observe that baseline and Fast-BEV achieve the best 33.8\% NDS and 45.1\% NDS at epoch 24 and 48, respectively. We find that the upper-bound of Fast-BEV is much higher than the baseline.
And Fast-BEV requires more training epochs to achieve better results, because of the strong data augmentation and temporal feature fusion.

\section{Conclusions}
In this paper, we propose Fast-BEV, a stronger and faster fully convolutional BEV perception framework which is suitable for on-vehicle depolyment.
Compared with the M$^2$BEV baseline, the new Fast-BEV introduces the strong 3D augmentation and temporal feature information, which largely boosts up the performance.
We also optimize the view transformation to make it more deployment friendly.
We propose to pre-compute the projection index and let all camera project to the same dense voxel.
We hope our work can shed light on the industrial-level, real-time, on-vehicle BEV perception.


\begin{thebibliography}{10}

\bibitem{liu2022bevfusion}
Zhijian Liu, Haotian Tang, Alexander Amini, Xinyu Yang, Huizi Mao, Daniela Rus,
  and Song Han.
\newblock Bevfusion: Multi-task multi-sensor fusion with unified bird's-eye
  view representation.
\newblock {\em arXiv preprint arXiv:2205.13542}, 2022.

\bibitem{geiger2012we}
Andreas Geiger, Philip Lenz, and Raquel Urtasun.
\newblock Are we ready for autonomous driving? the kitti vision benchmark
  suite.
\newblock In {\em 2012 IEEE conference on computer vision and pattern
  recognition}, pages 3354--3361. IEEE, 2012.

\bibitem{zhu2019class}
Benjin Zhu, Zhengkai Jiang, Xiangxin Zhou, Zeming Li, and Gang Yu.
\newblock Class-balanced grouping and sampling for point cloud 3d object
  detection.
\newblock {\em arXiv preprint arXiv:1908.09492}, 2019.

\bibitem{caesar2020nuscenes}
Holger Caesar, Varun Bankiti, Alex~H Lang, Sourabh Vora, Venice~Erin Liong,
  Qiang Xu, Anush Krishnan, Yu~Pan, Giancarlo Baldan, and Oscar Beijbom.
\newblock nuscenes: A multimodal dataset for autonomous driving.
\newblock In {\em Proceedings of the IEEE/CVF conference on computer vision and
  pattern recognition}, pages 11621--11631, 2020.

\bibitem{sun2020scalability}
Pei Sun, Henrik Kretzschmar, Xerxes Dotiwalla, Aurelien Chouard, Vijaysai
  Patnaik, Paul Tsui, James Guo, Yin Zhou, Yuning Chai, Benjamin Caine, et~al.
\newblock Scalability in perception for autonomous driving: Waymo open dataset.
\newblock In {\em Proceedings of the IEEE/CVF conference on computer vision and
  pattern recognition}, pages 2446--2454, 2020.

\bibitem{brazil2019m3d}
Garrick Brazil and Xiaoming Liu.
\newblock M3d-rpn: Monocular 3d region proposal network for object detection.
\newblock In {\em Proceedings of the IEEE/CVF International Conference on
  Computer Vision}, pages 9287--9296, 2019.

\bibitem{wang2021fcos3d}
Tai Wang, Xinge Zhu, Jiangmiao Pang, and Dahua Lin.
\newblock Fcos3d: Fully convolutional one-stage monocular 3d object detection.
\newblock In {\em Proceedings of the IEEE/CVF International Conference on
  Computer Vision}, pages 913--922, 2021.

\bibitem{tian2019fcos}
Zhi Tian, Chunhua Shen, Hao Chen, and Tong He.
\newblock Fcos: Fully convolutional one-stage object detection.
\newblock In {\em Proceedings of the IEEE/CVF international conference on
  computer vision}, pages 9627--9636, 2019.

\bibitem{wang2022probabilistic}
Tai Wang, ZHU Xinge, Jiangmiao Pang, and Dahua Lin.
\newblock Probabilistic and geometric depth: Detecting objects in perspective.
\newblock In {\em Conference on Robot Learning}, pages 1475--1485. PMLR, 2022.

\bibitem{park2021pseudo}
Dennis Park, Rares Ambrus, Vitor Guizilini, Jie Li, and Adrien Gaidon.
\newblock Is pseudo-lidar needed for monocular 3d object detection?
\newblock In {\em Proceedings of the IEEE/CVF International Conference on
  Computer Vision}, pages 3142--3152, 2021.

\bibitem{wang2019pseudo}
Yan Wang, Wei-Lun Chao, Divyansh Garg, Bharath Hariharan, Mark Campbell, and
  Kilian~Q Weinberger.
\newblock Pseudo-lidar from visual depth estimation: Bridging the gap in 3d
  object detection for autonomous driving.
\newblock In {\em Proceedings of the IEEE/CVF Conference on Computer Vision and
  Pattern Recognition}, pages 8445--8453, 2019.

\bibitem{you2019pseudo}
Yurong You, Yan Wang, Wei-Lun Chao, Divyansh Garg, Geoff Pleiss, Bharath
  Hariharan, Mark Campbell, and Kilian~Q Weinberger.
\newblock Pseudo-lidar++: Accurate depth for 3d object detection in autonomous
  driving.
\newblock {\em arXiv preprint arXiv:1906.06310}, 2019.

\bibitem{qian2020end}
Rui Qian, Divyansh Garg, Yan Wang, Yurong You, Serge Belongie, Bharath
  Hariharan, Mark Campbell, Kilian~Q Weinberger, and Wei-Lun Chao.
\newblock End-to-end pseudo-lidar for image-based 3d object detection.
\newblock In {\em Proceedings of the IEEE/CVF Conference on Computer Vision and
  Pattern Recognition}, pages 5881--5890, 2020.

\bibitem{rukhovich2022imvoxelnet}
Danila Rukhovich, Anna Vorontsova, and Anton Konushin.
\newblock Imvoxelnet: Image to voxels projection for monocular and multi-view
  general-purpose 3d object detection.
\newblock In {\em Proceedings of the IEEE/CVF Winter Conference on Applications
  of Computer Vision}, pages 2397--2406, 2022.

\bibitem{reading2021categorical}
Cody Reading, Ali Harakeh, Julia Chae, and Steven~L Waslander.
\newblock Categorical depth distribution network for monocular 3d object
  detection.
\newblock In {\em Proceedings of the IEEE/CVF Conference on Computer Vision and
  Pattern Recognition}, pages 8555--8564, 2021.

\bibitem{philion2020lift}
Jonah Philion and Sanja Fidler.
\newblock Lift, splat, shoot: Encoding images from arbitrary camera rigs by
  implicitly unprojecting to 3d.
\newblock In {\em European Conference on Computer Vision}, pages 194--210.
  Springer, 2020.

\bibitem{roddick2018orthographic}
Thomas Roddick, Alex Kendall, and Roberto Cipolla.
\newblock Orthographic feature transform for monocular 3d object detection.
\newblock {\em arXiv preprint arXiv:1811.08188}, 2018.

\bibitem{huang2021bevdet}
Junjie Huang, Guan Huang, Zheng Zhu, and Dalong Du.
\newblock Bevdet: High-performance multi-camera 3d object detection in
  bird-eye-view.
\newblock {\em arXiv preprint arXiv:2112.11790}, 2021.

\bibitem{xie2022m}
Enze Xie, Zhiding Yu, Daquan Zhou, Jonah Philion, Anima Anandkumar, Sanja
  Fidler, Ping Luo, and Jose~M Alvarez.
\newblock M\^{} 2bev: Multi-camera joint 3d detection and segmentation with
  unified birds-eye view representation.
\newblock {\em arXiv preprint arXiv:2204.05088}, 2022.

\bibitem{li2022bevformer}
Zhiqi Li, Wenhai Wang, Hongyang Li, Enze Xie, Chonghao Sima, Tong Lu, Qiao Yu,
  and Jifeng Dai.
\newblock Bevformer: Learning bird's-eye-view representation from multi-camera
  images via spatiotemporal transformers.
\newblock {\em arXiv preprint arXiv:2203.17270}, 2022.

\bibitem{carion2020end}
Nicolas Carion, Francisco Massa, Gabriel Synnaeve, Nicolas Usunier, Alexander
  Kirillov, and Sergey Zagoruyko.
\newblock End-to-end object detection with transformers.
\newblock In {\em European conference on computer vision}, pages 213--229.
  Springer, 2020.

\bibitem{wang2022detr3d}
Yue Wang, Vitor~Campagnolo Guizilini, Tianyuan Zhang, Yilun Wang, Hang Zhao,
  and Justin Solomon.
\newblock Detr3d: 3d object detection from multi-view images via 3d-to-2d
  queries.
\newblock In {\em Conference on Robot Learning}, pages 180--191. PMLR, 2022.

\bibitem{chen2022graph}
Zehui Chen, Zhenyu Li, Shiquan Zhang, Liangji Fang, Qinhong Jiang, and Feng
  Zhao.
\newblock Graph-detr3d: Rethinking overlapping regions for multi-view 3d object
  detection.
\newblock {\em arXiv preprint arXiv:2204.11582}, 2022.

\bibitem{yin2021center}
Tianwei Yin, Xingyi Zhou, and Philipp Krahenbuhl.
\newblock Center-based 3d object detection and tracking.
\newblock In {\em Proceedings of the IEEE/CVF conference on computer vision and
  pattern recognition}, pages 11784--11793, 2021.

\bibitem{lang2019pointpillars}
Alex~H Lang, Sourabh Vora, Holger Caesar, Lubing Zhou, Jiong Yang, and Oscar
  Beijbom.
\newblock Pointpillars: Fast encoders for object detection from point clouds.
\newblock In {\em Proceedings of the IEEE/CVF conference on computer vision and
  pattern recognition}, pages 12697--12705, 2019.

\bibitem{yan2018second}
Yan Yan, Yuxing Mao, and Bo~Li.
\newblock Second: Sparsely embedded convolutional detection.
\newblock {\em Sensors}, 18(10):3337, 2018.

\bibitem{nabati2021centerfusion}
Ramin Nabati and Hairong Qi.
\newblock Centerfusion: Center-based radar and camera fusion for 3d object
  detection.
\newblock In {\em Proceedings of the IEEE/CVF Winter Conference on Applications
  of Computer Vision}, pages 1527--1536, 2021.

\bibitem{hu2021fiery}
Anthony Hu, Zak Murez, Nikhil Mohan, Sof{\'\i}a Dudas, Jeffrey Hawke, Vijay
  Badrinarayanan, Roberto Cipolla, and Alex Kendall.
\newblock Fiery: Future instance prediction in bird's-eye view from surround
  monocular cameras.
\newblock In {\em Proceedings of the IEEE/CVF International Conference on
  Computer Vision}, pages 15273--15282, 2021.

\bibitem{saha2021translating}
Avishkar Saha, Oscar~Mendez Maldonado, Chris Russell, and Richard Bowden.
\newblock Translating images into maps.
\newblock {\em arXiv preprint arXiv:2110.00966}, 2021.

\bibitem{can2020understanding}
Yigit~Baran Can, Alexander Liniger, Ozan Unal, Danda Paudel, and Luc Van~Gool.
\newblock Understanding bird's-eye view semantic hd-maps using an onboard
  monocular camera.
\newblock {\em arXiv preprint arXiv:2012.03040}, 2020.

\bibitem{huang2022bevdet4d}
Junjie Huang and Guan Huang.
\newblock Bevdet4d: Exploit temporal cues in multi-camera 3d object detection.
\newblock {\em arXiv preprint arXiv:2203.17054}, 2022.

\bibitem{liu2022petr}
Yingfei Liu, Tiancai Wang, Xiangyu Zhang, and Jian Sun.
\newblock Petr: Position embedding transformation for multi-view 3d object
  detection.
\newblock {\em arXiv preprint arXiv:2203.05625}, 2022.

\bibitem{liu2022petrv2}
Yingfei Liu, Junjie Yan, Fan Jia, Shuailin Li, Qi~Gao, Tiancai Wang, Xiangyu
  Zhang, and Jian Sun.
\newblock Petrv2: A unified framework for 3d perception from multi-camera
  images.
\newblock {\em arXiv preprint arXiv:2206.01256}, 2022.

\bibitem{pan2020cross}
Bowen Pan, Jiankai Sun, Ho~Yin~Tiga Leung, Alex Andonian, and Bolei Zhou.
\newblock Cross-view semantic segmentation for sensing surroundings.
\newblock {\em IEEE Robotics and Automation Letters}, 5(3):4867--4873, 2020.

\bibitem{roddick2020predicting}
Thomas Roddick and Roberto Cipolla.
\newblock Predicting semantic map representations from images using pyramid
  occupancy networks.
\newblock In {\em Proceedings of the IEEE/CVF Conference on Computer Vision and
  Pattern Recognition}, pages 11138--11147, 2020.

\bibitem{yang2021projecting}
Weixiang Yang, Qi~Li, Wenxi Liu, Yuanlong Yu, Yuexin Ma, Shengfeng He, and Jia
  Pan.
\newblock Projecting your view attentively: Monocular road scene layout
  estimation via cross-view transformation.
\newblock In {\em Proceedings of the IEEE/CVF Conference on Computer Vision and
  Pattern Recognition}, pages 15536--15545, 2021.

\bibitem{li2022bevdepth}
Yinhao Li, Zheng Ge, Guanyi Yu, Jinrong Yang, Zengran Wang, Yukang Shi,
  Jianjian Sun, and Zeming Li.
\newblock Bevdepth: Acquisition of reliable depth for multi-view 3d object
  detection.
\newblock {\em arXiv preprint arXiv:2206.10092}, 2022.

\bibitem{milioto2019rangenet++}
Andres Milioto, Ignacio Vizzo, Jens Behley, and Cyrill Stachniss.
\newblock Rangenet++: Fast and accurate lidar semantic segmentation.
\newblock In {\em 2019 IEEE/RSJ international conference on intelligent robots
  and systems (IROS)}, pages 4213--4220. IEEE, 2019.

\bibitem{zhang2022beverse}
Yunpeng Zhang, Zheng Zhu, Wenzhao Zheng, Junjie Huang, Guan Huang, Jie Zhou,
  and Jiwen Lu.
\newblock Beverse: Unified perception and prediction in birds-eye-view for
  vision-centric autonomous driving.
\newblock {\em arXiv preprint arXiv:2205.09743}, 2022.

\bibitem{chitta2021neat}
Kashyap Chitta, Aditya Prakash, and Andreas Geiger.
\newblock Neat: Neural attention fields for end-to-end autonomous driving.
\newblock In {\em Proceedings of the IEEE/CVF International Conference on
  Computer Vision}, pages 15793--15803, 2021.

\bibitem{gosala2022bird}
Nikhil Gosala and Abhinav Valada.
\newblock Bird’s-eye-view panoptic segmentation using monocular frontal view
  images.
\newblock {\em IEEE Robotics and Automation Letters}, 7(2):1968--1975, 2022.

\bibitem{zhou2019objects}
Xingyi Zhou, Dequan Wang, and Philipp Kr{\"a}henb{\"u}hl.
\newblock Objects as points.
\newblock {\em arXiv preprint arXiv:1904.07850}, 2019.

\bibitem{lu2022learning}
Jiachen Lu, Zheyuan Zhou, Xiatian Zhu, Hang Xu, and Li~Zhang.
\newblock Learning ego 3d representation as ray tracing.
\newblock {\em arXiv preprint arXiv:2206.04042}, 2022.

\bibitem{jiang2022polarformer}
Yanqin Jiang, Li~Zhang, Zhenwei Miao, Xiatian Zhu, Jin Gao, Weiming Hu, and
  Yu-Gang Jiang.
\newblock Polarformer: Multi-camera 3d object detection with polar
  transformers.
\newblock {\em arXiv preprint arXiv:2206.15398}, 2022.

\bibitem{zhou2018voxelnet}
Yin Zhou and Oncel Tuzel.
\newblock Voxelnet: End-to-end learning for point cloud based 3d object
  detection.
\newblock In {\em Proceedings of the IEEE conference on computer vision and
  pattern recognition}, pages 4490--4499, 2018.

\bibitem{shi2019pointrcnn}
Shaoshuai Shi, Xiaogang Wang, and Hongsheng Li.
\newblock Pointrcnn: 3d object proposal generation and detection from point
  cloud.
\newblock In {\em Proceedings of the IEEE/CVF conference on computer vision and
  pattern recognition}, pages 770--779, 2019.

\bibitem{neuvition_2022}
Lidar price for cars - neuvition: Solid-state lidar,lidar sensor suppliers,
  lidar technology, lidar sensor, Feb 2022.

\end{thebibliography}


\end{document}